\documentclass{SCIS2019}

\begin{document}
\ArticleType{LETTER}
\Year{2019}
\Month{}
\Vol{}
\No{}
\DOI{}
\ArtNo{}
\ReceiveDate{}
\ReviseDate{}
\AcceptDate{}
\OnlineDate{}

\title{MDSSD: Multi-scale Deconvolutional Single\\ Shot Detector for Small Objects}{MDSSD: Multi-scale Deconvolutional Single Shot Detector for Small Objects}


\author[]{Lisha CUI}{}
\author[]{Rui MA}{maruifirst@zzu.edu.cn}
\author[]{Pei LV}{}
\author[]{Xiaoheng JIANG}{}
\author[]{Zhimin GAO}{}
\author[]{Bing ZHOU}{}
\author[]{Mingliang XU}{}

\AuthorMark{CUI L S, et al}

\AuthorCitation{CUI L S, MA R, LV P, et al}



\address[]{Center for Interdisciplinary Information Science Research, Zhengzhou University, Zhengzhou 450001, China}

\maketitle

\begin{multicols}{2}
\deareditor
Object detection has always been the focus and challenge in the field of computer vision, especially the small object detection. With the arrival of deep convolutional networks (ConvNets~\cite{CNN}), the performance of object detection \cite{Faster,SSD,FPN,couple} has been improved significantly. However, small object detection is still a challenging issue due to the relatively small area with less information in images.

Recent detection systems such as Faster R-CNN \cite{Faster} leverage the top-most feature maps within ConvNets on a single input scale to predict candidate bounding boxes of different scales and aspect ratios. However, the top-most feature maps have a fixed receptive field, which conflicts with objects at different scales in natural images. In addition, there is little information left on the top-most feature maps for small objects, which compromises the detection performance, especially for small object detection.

To address the problem of multi-scale object detection, the methods such as SSD \cite{SSD} utilize the pyramidal feature hierarchy from bottom to top layers to detect objects of various sizes. Nevertheless, the features from the bottom layers of ConvNets have weak semantic information, which could harm their representational capacity for small object recognition.

The most recent networks such as FPN~\cite{FPN} try to make full use of the pyramidal features by building a top-down architecture with lateral connections. These networks show dramatic improvements in accuracy compared with conventional detectors. These systems, however, first progressively reduce the input image to small feature maps which retain little spatial information of small objects, and then try to reconstruct the spatial resolution. In fact, it is difficult to restore the lost spatial information of small objects by upsampling.
More details of the related work are given in Appendix A.

In this study, we design a Multi-scale Deconvolutional Single Shot Detector (MDSSD), especially for small object detection. In MDSSD, multiple high-level feature maps at different scales are upsampled simultaneously to increase the spatial resolution. Afterwards, we implement the skip connections with low-level feature maps via a Fusion Block. The fusion feature maps, named Fusion Module, are of strong feature representation power of small instances. It is noteworthy that these high-level feature maps utilized in the Fusion Block preserve both strong semantic information and some fine details of small instances, unlike the top-most layer where the representation of fine details for small objects is potentially wiped out.

The main contributions of our work are summarized as follows:

\begin{figure*}[t]
\centering
\includegraphics[width=12cm]{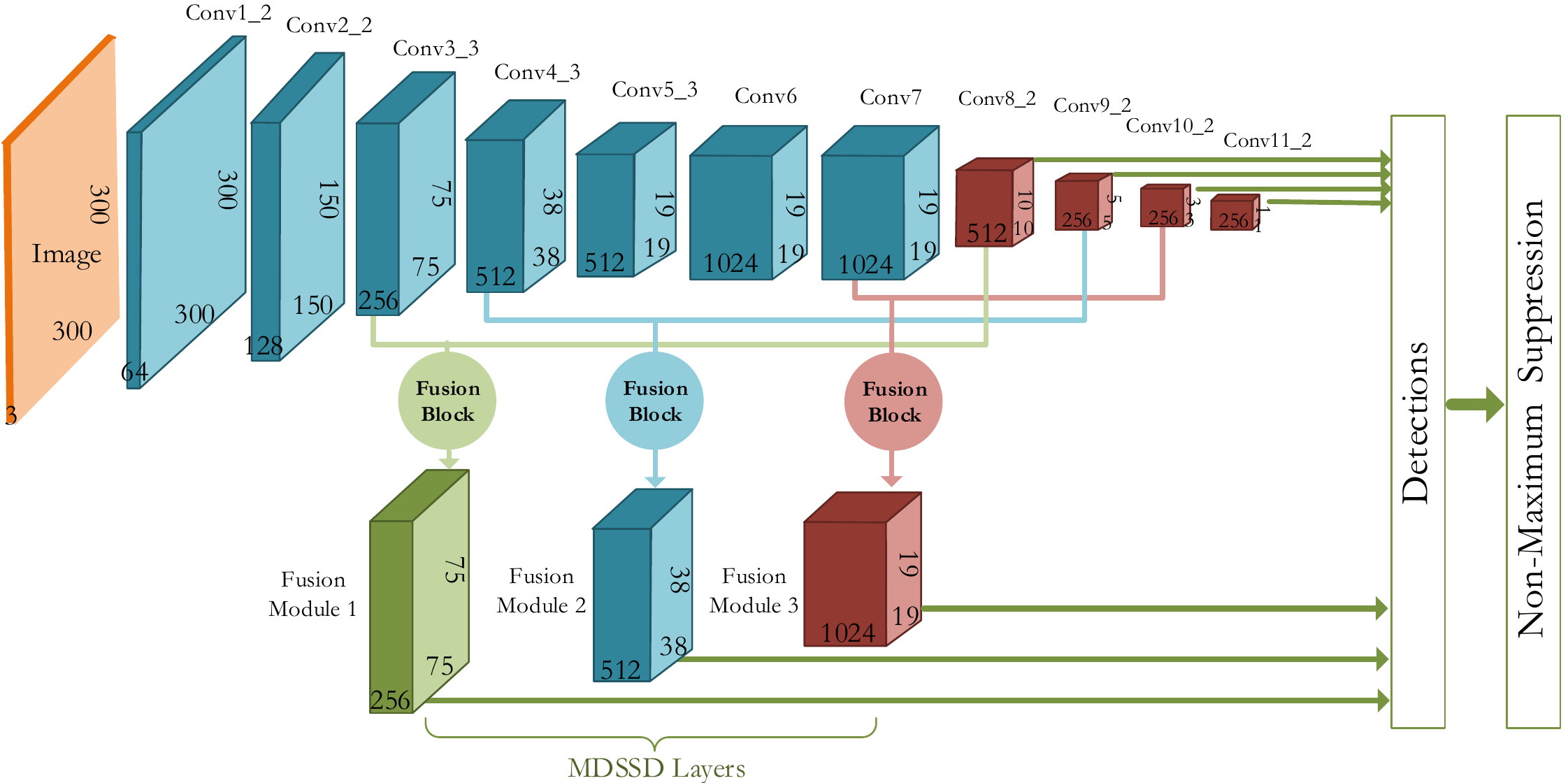}
\caption{The architecture of MDSSD. First, we apply deconvolution layers to the high-level semantic feature maps at different scales (i.e., conv8\_2, conv9\_2, and conv$10\_2$) simultaneously. Then we build skip connections with lower-layers (conv3\_3, conv4\_3, and conv7) through Fusion Block and form 3 new fusion layers (Module 1, Module 2, and Module 3). Predictions are made on both new fusion layers (Module 1, Module 2, and Module 3) and original SSD layers (conv8\_2, conv9\_2, conv$10\_2$, and conv11\_2) at the same time.}
\label{architecture}
\end{figure*}

 (1) We design the delicate Fusion Block to build high-level semantic feature maps at different scales. The new Fusion Modules have both strong semantic information and accurate location information for small object detection.

 (2) We propose a novel framework MDSSD for small object detection. Several Fusion Blocks are applied on multi-scale convolution layers before the top-most feature maps where fine details still exist, providing a significant improvement in small object detection.


\lettersection{Methodology}
To figure out the effect of feature resolution on small object detection, we conduct pilot experiments on the benchmark TT100K \cite{TT100K} with SSD model.
First, we fine-tune the overall SSD512 and SSD300 models and test each prediction layer separately. Second, we fine-tune each prediction layer of SSD model independently and test each corresponding layer. The experiments show that the mAPs of these prediction layers (from conv4 to conv12) progressively drop to zero for both SSD512 and SSD300 models. That is to say, the representational capacity of small objects becomes worse and worse along with the depth of network increasing until totally lost on the coarse-semantic deeper layers (after conv11). The details are given in Appendix B.

(1) \emph{The Architecture of MDSSD.}
Based on the aforementioned observation, we devise the overall framework of MDSSD, as shown in Figure \ref{architecture}. To make the feature maps of the shallow layers contain more semantic information, we build several fusion layers between high-level and low-level feature maps through Fusion Blocks. As the pilot experiments analyzed, conv11 and conv12 have totally lost the fine details of small objects, and therefore we apply the Fusion Block before conv11. In order to share the structure of Fusion Block, we delicately design \emph{symmetric} topology between shallow layers and deep layers. That is to say, these shallow feature maps should have the same downsampling factor with the corresponding deep feature maps in terms of spatial resolution.

Specifically, conv4\_3 and conv7 are merged with conv9\_2 and conv10\_2 through Fusion Blocks, respectively. The new fusion feature maps, termed Fusion Module 2 and Fusion Module 3, are used to replace the original conv4\_3 and conv7 of SSD for detection. In order to further improve the performance of small object detection, it is necessary to take full advantage of the shallow feature maps. Therefore, we add Fusion Module 1 which connects the lower layer conv3\_3 and layer conv8\_2 to make predictions. The Fusion Modules can capture both more fine details and strong semantic information of small instances, and thus significantly improve the performance. Note that the three pairs of fusion layers (conv3\_3 and conv8\_2, conv4\_3 and conv9\_2, conv10\_2 and conv7) undergo downsampling by the same factor of 8; therefore they can share the same structure of Fusion Block.

In summary, we have seven prediction layers at different depths in total, including three Fusion Modules (Module 1, Module 2, and Module 3) and four original SSD prediction layers (conv8\_2, conv9\_2, comv10\_2, and conv11\_2). Fusion Modules are mainly responsible for accurately detecting relatively small instances, while the rest of the layers for detecting medium and large objects.

(2) \emph{Fusion Block.} There are three Fusion Modules at different depths. We take Module 1 as an example for interpretation, and the details of the architecture of Fusion Block are shown in Appendix B for the 300$\times$300 input model. The feature maps should have the same size and channels if we use the element-wise product or summation to merge them together. Therefore, in order to fuse conv3\_3 and conv8\_2, we need to upsample the spatial resolution of conv8\_2 by a factor of $8$.

Specifically, for conv8\_2 with the size of 10 $\times$ 10, we implement three deconvolution layers to achieve upsampling, with each upsampling the input feature maps by stride 2. The kernel size of the deconvolution layer is $2\times2$ or $3\times3$ with 256 channels. Each deconvolution layer is followed by one convolution layer. The resolution of feature maps produced by the upper branch is 75 $\times$ 75 after one L2 normalization layer. Conv3\_3 undergoes one 3$\times$3 convolution layer with stride 1 and L2 normalization. We fuse the feature maps output by the upper and lower branches through element-wise summation. Then we add one convolution layer to ensure the discriminability of features for detection. Finally, we obtain the Fusion Module 1 after a ReLU activation function.

As mentioned before, the \emph{symmetric} connections enable Fusion Module 2 and Module 3 to follow the identical structure. The only difference is that the numbers of channels of the three modules are 256, 512 and 1024, respectively. As for the 512$\times$512 input model, there are some tiny modifications. Further details of the Fusion Block with 300$\times$300 and 512$\times$512 input are provided in Appendix B.
The training objective is the weighted sum between localization loss (Smooth L1) and confidence loss (Softmax). More details of the training strategies are given in Appendix B.

\lettersection{Experiments}
We evaluate MDSSD on small object dataset TT100K \cite{TT100K}, the benchmark datasets PASCAL VOC2007~\cite{VOC} and MS COCO~\cite{COCO}.

(1) \emph{Results on TT100K.} MDSSD512 achieves an mAP of 77.6\% on TT100K, outperforming SSD512 (68.7\%) and RFB Net (74.4\%) by 8.9 and 3.2 points respectively with the same input size (512 $\times$ 512) and backbone. Moreover, MDSSD512 also exceeds the two-stage detector Faster R-CNN (52.9\% (VGG-based) and 61.1\% (ResNet-based)) with a large margin, even though they have a larger input size of $1000 \times 600$. Besides, FPN (69.9\%) and Mask R-CNN (70.8\%), the variants of Faster R-CNN, are also inferior to the proposed model. The results on TT100K demonstrate the effectiveness of MDSSD for small object detection.

(2) \emph{Results on PASCAL VOC2007.} MDSSD300 (78.6\%) exceeds the latest SSD300* (77.5\%) and is comparable to DSSD321 (78.6\%) on VOC2007.
Additionally, by replacing the backbone with ResNet-101, MDSSD320* (79.1\%) and MDSSD512* (81.0\%) achieve better results than the original models. Note that, MDSSD320* yields 0.5 points gain compared with DSSD321, while MDSSD512* is slightly inferior to DSSD513. The results demonstrate that MDSSD gains greater improvements for small input size.

(3) \emph{Results on COCO.}
It is noticeable that the proposed MDSSD300 and MDSSD512 models achieve 10.8\% AP and 13.9\% AP for small objects (area \textless   32$^2$), respectively. Our models outperform SSD (6.6\% and 10.9\%), DSSD (7.4\% and 13\%), and DSOD (9.4\%/-) with a large margin. MDSSD outperforms all one-stage architectures based on both VGG16 and ResNet-101. The proposed method achieves a higher AR (average recall) for small objects as well, which proves that MDSSD is more powerful for small object detection.
The details of the experimental results are provided in Appendix C.

\lettersection{Conclusions}
This paper proposes a Multi-scale Deconvolutional Single Shot Detector for small objects. We devise several Fusion Modules having different spatial resolutions to better match small objects. The skip connections add context information to low-level feature maps and make them more descriptive. While we only take SSD as the base architecture for demonstration, the principle can be also applied to other object detectors, such as Faster R-CNN~\cite{Faster}.

\Acknowledgements{This work was supported by National Natural Science Foundation of China (Grant Nos. 61822701, 61672469, 61772474, 61802351, 61872324).}


\Supplements{Appendix A, B, and C.}


\end{multicols}

\begin{thebibliography}{99}


\bibitem{CNN} Krizhevsky A, Sutskever I, Hinton G E. Imagenet classification with deep convolutional neural networks. In: Advances in Neural Information Processing Systems, Lake Tahoe, 2012. 1097--1105


\bibitem{Faster} Ren S Q, He K M, Girshick R B, et al. Faster r-cnn: towards real-time object detection with region proposal networks. In: Advances in Neural Information Processing Systems, Montreal, 2015. 91--99



\bibitem{SSD} Liu W, Anguelov D, Erhan D, et al. Ssd: Single shot multibox detector. In: European Conference on Computer Vision, Amsterdam, 2016. 21--37



\bibitem{FPN} Lin T Y, Doll{\'{a}}r P, Girshick R B, et al. Feature pyramid networks for object detection. In: Computer Vision and Pattern Recognition, Honolulu, 2017. 936--944

\bibitem{couple} Zhao J P, Guo W W, Zhang Z H, et al. A coupled convolutional neural network for small and densely clustered ship detection in SAR images. Sci China Inf Sci, 2019, 62: 042301



\bibitem{TT100K} Zhu Z, Liang D, Zhang S H, et al. Traffic-sign detection and classification in the wild. In: Computer Vision and Pattern Recognition, Las Vegas, 2016. 2110--2118


\bibitem{VOC} Everingham M, Gool L V, Williams C K I, et al. The pascal visual object classes (voc) challenge. Int J Comput Vision, 2010, 88: 303--338


\bibitem{COCO} Lin T Y, Maire M, Belongie S J, et al. Microsoft coco: Common objects in context. In: European Conference on Computer Vision, Zurich, 2014. 740--755

\end{thebibliography}
\end{document}